\documentclass[]{article}
\usepackage{arxiv}
\usepackage{cite}
\usepackage{amsmath,amssymb,amsfonts}
\usepackage{algorithmic}
\usepackage{graphicx}
\usepackage{textcomp}
\def\BibTeX{{\rm B\kern-.05em{\sc i\kern-.025em b}\kern-.08em
    T\kern-.1667em\lower.7ex\hbox{E}\kern-.125emX}}

\usepackage{siunitx}
\usepackage{url}
\usepackage{hyperref}
\usepackage{doi}
\usepackage[labelfont=bf, up]{caption}

\title{Leveraging Swin Transformer for enhanced diagnosis of Alzheimer’s disease using multi-shell diffusion MRI}

\usepackage{authblk}

\newbox{\orcid}\sbox{\orcid}{\includegraphics[scale=0.06]{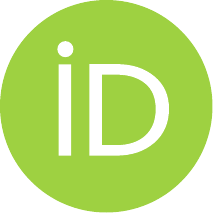}} 
\author[1,2]{%
	\href{https://orcid.org/0000-0002-7314-0413}{\usebox{\orcid}\hspace{1mm}Quentin Dessain\thanks{\texttt{quentin.dessain@uclouvain.be}}}%
}
\author[1,2]{%
	\href{https://orcid.org/0000-0001-6783-6601}{\usebox{\orcid}\hspace{1mm}Nicolas Delinte}%
}
\author[2]{%
	\href{https://orcid.org/0000-0002-3102-6778}{\usebox{\orcid}\hspace{1mm}Bernard Hanseeuw}%
}
\author[2]{%
	\href{https://orcid.org/0000-0002-8613-2420}{\usebox{\orcid}\hspace{1mm}Laurence Dricot}%
}
\author[1]{%
	\href{https://orcid.org/0000-0002-7243-4778}{\usebox{\orcid}\hspace{1mm}Benoît Macq}%
}
\affil[1]{ICTEAM Institute, UCLouvain, 1348 Louvain-la-Neuve, Belgium}
\affil[2]{Institute of Neuroscience (IoNS), UCLouvain, 1200 Brussels, Belgium}

\begin{document}
\maketitle

\begin{abstract}
\textit{Objective:} This study aims to support early diagnosis of Alzheimer’s disease and detection of amyloid accumulation by leveraging the microstructural information available in multi-shell diffusion MRI (dMRI) data, using a vision transformer-based deep learning framework.\\

\textit{Methods:} We present a classification pipeline that employs the Swin Transformer, a hierarchical vision transformer model, on multi-shell dMRI data for the classification of Alzheimer’s disease and amyloid presence. Key metrics from DTI and NODDI were extracted and projected onto 2D planes to enable transfer learning with ImageNet-pretrained models. To efficiently adapt the transformer to limited labeled neuroimaging data, we integrated Low-Rank Adaptation. We assessed the framework on diagnostic group prediction (cognitively normal, mild cognitive impairment, Alzheimer's disease dementia) and amyloid status classification.\\

\textit{Results:} The framework achieved competitive classification results within the scope of multi-shell dMRI-based features, with the best balanced accuracy of 95.2\% for distinguishing cognitively normal individuals from those with Alzheimer’s disease dementia using NODDI metrics. For amyloid detection, it reached 77.2\% balanced accuracy in distinguishing amyloid-positive mild cognitive impairment/Alzheimer’s disease dementia subjects from amyloid-negative cognitively normal subjects, and 67.9\%  for identifying amyloid-positive individuals among cognitively normal subjects. Grad-CAM-based explainability analysis identified clinically relevant brain regions, including the parahippocampal gyrus and hippocampus, as key contributors to model predictions.\\

\textit{Conclusion/Significance:} This study demonstrates the promise of diffusion MRI and transformer-based architectures for early detection of Alzheimer’s disease and amyloid pathology, supporting biomarker-driven diagnostics in data-limited biomedical settings.
\end{abstract}

\keywords{Alzheimer's disease \and deep learning \and diffusion MRI \and DTI \and NODDI \and Swin Transformer}

\begin{center}
    \textit{This work has been submitted to the IEEE for possible publication. Copyright may be transferred without notice, after which this version may no longer be accessible.}
\end{center}

\clearpage

\section{Introduction}
\label{sec:introduction}

Diffusion Magnetic Resonance Imaging (dMRI) has emerged as a powerful modality to investigate the microstructural properties of brain tissue. By analyzing the diffusion of water molecules in the brain, dMRI provides insights into the integrity of white matter tracts and the microstructural organization of brain tissue. In particular, the combination of high angular resolution diffusion imaging (HARDI) with multi-shell dMRI offers a detailed characterization of tissue microstructure, positioning it as a promising tool for diagnosing and understanding the underlying biological changes associated with neurodegenerative diseases \cite{frank2001anisotropy}.

However, the high dimensionality and complexity of multi-shell dMRI data present significant challenges for traditional analysis methods. Deep learning has shown promise in automating feature extraction and classification from high-dimensional neuroimaging data \cite{litjens2017survey}. Convolutional Neural Networks (CNNs), in particular, have demonstrated substantial success in medical image analysis, including brain MRI tasks, due to their ability to capture local spatial features~\cite{lecun1998gradient}. These approaches have been successfully applied to structural MRI (sMRI) and functional MRI (fMRI) for Alzheimer's disease classification~\cite{wen2020convolutional}, with models such as hierarchical fully convolutional networks extracting multi-scale features from sMRI \cite{lianhierarchical2020} and spatiotemporal 3D CNNs capturing dynamics in fMRI data \cite{parmar2020spatiotemporal}. However, CNNs primarily focus on local spatial features, which may limit their ability to model global context and long-range dependencies.

To address these limitations, transformer models, originally developed for natural language processing tasks, have recently gained traction in medical imaging due to their ability to capture both local and global context through self-attention mechanisms. The Vision Transformer (ViT), which applies these principles to image data, has shown competitive performance across various computer vision tasks by leveraging its ability to model long-range dependencies in images~\cite{dosovitskiy_image_2021}. In the neuroimaging field, pre-trained ViTs have demonstrated promising results in tasks such as brain tumor classification and Alzheimer’s disease diagnosis using MRI~\cite{he2023transformers},~\cite{shamshad2023transformers},~\cite{alp_joint_2024}.
However, ViTs typically require large-scale labeled datasets for effective training, which can be a limiting factor in medical imaging where labeled data are often scarce~\cite{touvron2021training}. 
While using pre-trained ViT can alleviate the need for extensive labeled data, ViTs still face challenges in medical imaging due to their fixed input size. This constraint limits the usability of pre-trained models when the input dimensions of medical images differ from the size expected by the pre-trained ViT~\cite{liu2021swin}.

The Swin Transformer, a variant of the ViT, addresses the former challenges by introducing a hierarchical structure and a shifted window approach, allowing it to capture both fine-grained local features and broader global dependencies while supporting variable input sizes~\cite{liu2021swin}. This flexibility is particularly beneficial for medical imaging, where the scale and resolution of images can vary significantly, and the ability to leverage pre-trained weights is critical for maximizing performance in the context of limited data availability~\cite{kim2022transfer,morid2021scoping}. 

The Swin Transformer’s ability to maintain high-resolution representations while efficiently handling large images makes it an ideal choice for analyzing complex neuroimaging data, such as those derived from multi-shell dMRI. This ability to work with high-resolution representations, combined with the use of pre-trained weights, allows Swin Transformers to not only improve the adaptability of models to different image sizes and resolutions, but also enhances the potential for accurate diagnosis and prognosis of neuropathologies such as Alzheimer's disease by leveraging diffusion MRI data.

The recent inclusion of multi-shell diffusion MRI in the Alzheimer’s Disease Neuroimaging Initiative (ADNI) dataset presents a valuable opportunity for developing new diagnostic tools~\cite{zavaliangos-petropulu_diffusion_2019}. Historically, most large diffusion MRI datasets were limited to single-shell acquisitions~\cite{mueller_ways_2005}, suitable primarily for Apparent Diffusion Coefficient (ADC) and Diffusion Tensor Imaging (DTI) \cite{basser_mr_1994}. While foundational, these models are constrained by simplified assumptions about water diffusion, limiting their ability to characterize complex microstructural features such as crossing fibers and tissue heterogeneity. Multi-shell dMRI overcomes these limitations by enabling more sophisticated models that provide deeper insights into tissue integrity and pathology, making it particularly valuable for studying Alzheimer’s disease, since early diagnosis and monitoring of Alzheimer’s disease are crucial for effective disease management, with the potential to delay progression and improve the quality of life for affected individuals. However, despite significant advances in neuroimaging and clinical diagnostics using DTI~\cite{acosta-cabronero_diffusion_2012},~\cite{demirhan_feature_2015},~\cite{ahmed_recognition_2017},~\cite{marzban_alzheimers_2020},~\cite{lella_ensemble_2021}, identifying reliable biomarkers for the early stages of Alzheimer's remains a significant challenge. 

In this study, we propose an architecture that leverages a multi-shell dMRI dataset and the recent developments of vision-oriented transformer models. The imaging data are preprocessed, and microstructural models, such as DTI~\cite{basser_mr_1994} and Neurite Orientation Dispersion and Density Imaging (NODDI)~\cite{zhang_noddi_2012}, are derived. The 3D microstructural maps estimated with these models were projected onto 2D planes, enabling the Swin Transformer to effectively operate on these 2D projections while retaining the advantages of pre-trained weights. By employing this advanced architecture and utilizing dMRI's unique insights into microstructural changes, we aim to enhance the understanding and diagnosis of Alzheimer’s disease.

\subsection{Contributions}

\begin{itemize}
    \item We highlight the use of multi-shell diffusion MRI for Alzheimer’s disease classification, leveraging the microstructural information available from diffusion models such as DTI and NODDI.
    \item We demonstrate the potential of using multi-shell diffusion MRI for predicting amyloid beta status, a key biomarker for early detection of Alzheimer’s disease.
    \item To efficiently process diffusion MRI data, we project 3D microstructural maps into 2D representations, allowing the use of pre-trained Swin Transformer while maintaining essential spatial information.
    \item We incorporate LoRA, a parameter-efficient fine-tuning approach, into our deep learning framework, enhancing adaptability to neuroimaging tasks with limited labeled data while leveraging pre-trained ImageNet weights.   
    \item Through Grad-CAM-based explainability, we highlight the clinically relevant brain regions that contribute to model classification decisions, offering valuable interpretability for medical applications.
    \item Our framework is designed to be adaptable and extendable to other neuroimaging modalities and medical imaging tasks, paving the way for future applications beyond Alzheimer's disease.
\end{itemize}

\section{Materials and Methods}
\label{sec:methods}

\subsection{Dataset}

The neuroimaging data used in the preparation of this article were obtained from the ADNI database (\url{adni.loni.usc.edu}), specifically from the ADNI3 and ADNI4 phases. The ADNI was launched in 2003 as a public–private partnership led by Principal Investigator Michael W. Weiner, MD. The primary goal of ADNI has been to test whether serial MRI, Positron Emission Tomography (PET), other biological markers, and clinical and neuropsychological assessment can be combined to measure the progression of mild cognitive impairment and early Alzheimer’s disease.

For this study, we used both multi-shell dMRI data and T1-weighted sMRI data from the ADNI repository. The dataset includes scans from participants categorized into three diagnostic groups: Cognitively Normal (CN), Mild Cognitive Impairment (MCI), and Alzheimer's disease Dementia (AD).

To leverage metrics from multi-shell models, single-shell dMRI acquisitions present in the ADNI dataset were excluded. All scans were acquired using a 3.0 Tesla Siemens MAGNETOM Prisma/Prisma$^\text{fit}$ scanner with a 2D echo-planar sequence. The voxel size was 2.0 × 2.0 × 2.0 mm. The acquisition parameters included an echo time (TE) of 71.0 ms, a repetition time (TR) of 3400.0 ms, and a flip angle of 90 degrees. The diffusion-weighted imaging protocol had 6 gradient directions at $b=500$, 48 directions at $b=1000$  and 60 directions at $b=\SI{2000}{\micro\second\per\micro\meter\squared}$, along with 13 $b=0$ acquisitions, for a total of 127 measurements. The matrix size for the scans was 116 × 116 × 81 pixels.

T1-weighted sMRI scans using the MPRAGE sequence, acquired during the same session as the diffusion data, were included to provide anatomical context for the neuroimaging analysis. These MPRAGE scans were acquired in the sagittal plane using a 3D gradient-recalled inversion recovery sequence, with a voxel size of 1.0 × 1.0 × 1.0 mm. The acquisition parameters included a TE of 3.0 ms, a TR of 2300.0 ms, and an inversion time of 900.0 ms. The flip angle was set to 9 degrees, and the matrix size was 240 × 256 × 208 pixels.

\begin{table}[h!]
\centering
\caption{Demographic characteristics of selected participants from the ADNI dataset. When available, subjects were further subclassified by amyloid beta (A$\beta$) status into A$\beta^{-}$ and A$\beta^{+}$.}
\label{tab:demographics}
\begin{tabular}{lccc}
\hline
 & AD & MCI & CN \\ \hline
Number of subjects & 37 & 121 & 213 \\
Amyloid status (A$\beta^{-}$/A$\beta^{+}$) & 0/17 & 42/40 & 107/54 \\
Sex (Male/Female) & 17/20 & 65/56 & 72/141 \\
Age (Mean $\pm$ SD) & $77.2 \pm 6.9$ & $73.3 \pm 7.2$ & $72.9 \pm 8.7$ \\
\hline
\end{tabular}
\end{table}

A total of 371 participants were included after preprocessing, comprising 37 individuals diagnosed with AD, 121 with MCI, and 213 CN subjects. The demographic characteristics are summarized in Table~\ref{tab:demographics}.

\subsection{Processing the dMRI data}
The preprocessing of diffusion MRI data was performed using ElikoPy~\cite{quentin_dessain_hyedrynelikopy_2024}, incorporating a series of specialized algorithms to ensure high-quality data suitable for subsequent analysis. The preprocessing began with MP-PCA denoising~\cite{veraart2016denoising}, applied through MRtrix3’s dwidenoise function (version 3.0.3)~\cite{tournier2019mrtrix3}. Following denoising, FSL’s eddy tool (version 6.0.7)~\cite{andersson2016integrated} was employed for both head motion correction and eddy current correction. Additionally, b=0 reference images were processed alongside T1-weighted images using Synb0disco~\cite{schilling2019synthesized} to estimate the susceptibility-induced off-resonance field. These field maps were then integrated into the eddy correction process, ensuring that both eddy current and head motion were accurately corrected during the interpolation phase. Finally, post-eddy alignment of shells was performed to further refine the data, ensuring precise spatial consistency across all diffusion-weighted images.

After preprocessing, microstructural models were derived from the multi-shell dMRI data, and three key metrics were selected for each model as inputs to the Swin Transformer. For DTI, the metrics included Fractional Anisotropy (FA), which measures the degree of anisotropy in water diffusion; Axial Diffusivity (AxD), indicating water diffusion along the principal axis of the diffusion tensor; and Radial Diffusivity (RD), representing diffusion perpendicular to the principal axis. For NODDI, the selected metrics were the Orientation Dispersion Index (ODI), which quantifies the spread of neurite orientations; the Intra-neurite Volume Fraction ($f_{intra}$), reflecting the proportion of the diffusion MRI signal from neurites; and the Extra-neurite Volume Fraction ($f_{extra}$), representing the signal from the extracellular space. These metrics were chosen for their relevance in characterizing microstructural properties and providing rich input features for the transformer model. The 3D microstructural maps derived from the dMRI data were subsequently used as input for the deep learning models developed in this study.

\begin{figure*}[htbp]
\centerline{\includegraphics[width=\textwidth]{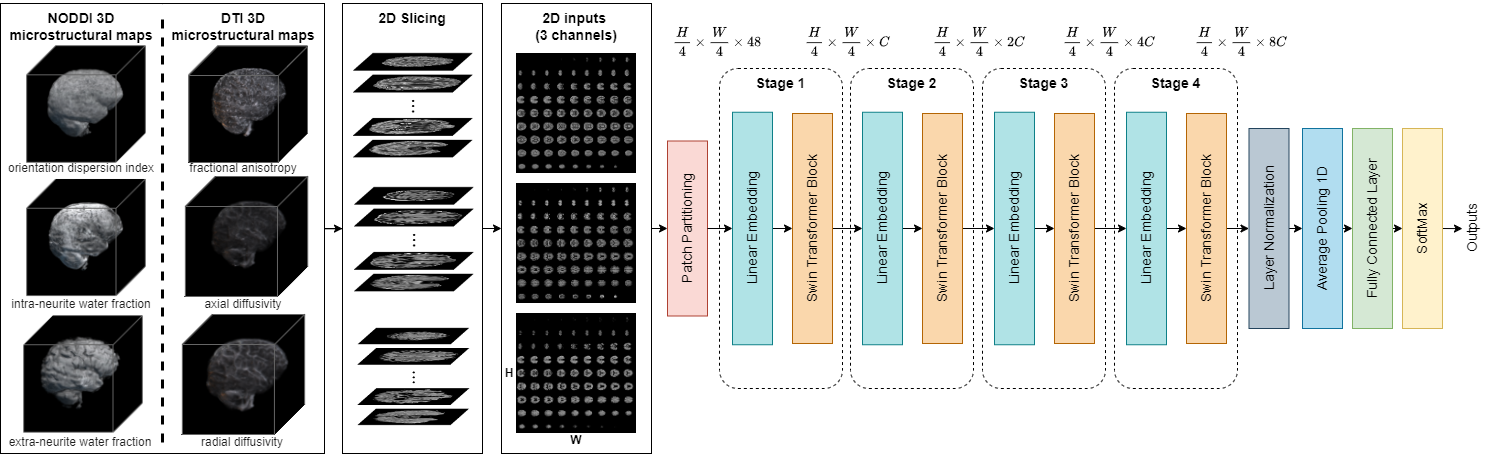}}
\caption{Overview of the proposed architecture. The 3D microstructural maps derived from diffusion MRI data are projected into 2D slices, which serve as input to the Swin Transformer pre-trained on ImageNet. The Swin Transformer blocks are fine-tuned with LoRA.}
\label{fig:swin}
\end{figure*}

\subsection{Network architecture}

For each model, the three selected 3D microstructural maps are projected into 2D space through orthogonal slicing, creating three-channel images where each channel corresponds to a specific microstructural map, as illustrated in Fig.~\ref{fig:swin}.

\begin{equation}
\mathbf{I}(x, y, c) = \mathbf{M}_c(x, y) \quad \text{for } c \in \{1, 2, 3\}.
\end{equation}

where \(\mathbf{I}(x, y, c)\) represents the resulting 2D image at coordinates \((x, y)\), with channel \(c \in \{1, 2, 3\}\) corresponding to one of the microstructural maps: \(\mathbf{M}_1(x, y)\), \(\mathbf{M}_2(x, y)\), or \(\mathbf{M}_3(x, y)\). These three-channel images are then input into a fine-tuned Swin Transformer architecture, pre-trained on the ImageNet dataset~\cite{deng2009imagenet}. 

To enhance the model's generalization across different native spaces and reduce overfitting to specific subject characteristics, we applied a data augmentation strategy to the 3D microstructural volumes before slicing them into 2D images. The augmentation technique consisted in synchronized translations by up to 5 voxels applied uniformly across the 3D volumes for all three channels, corresponding to the different microstructural maps. 

\begin{figure}[htbp]
\centerline{\includegraphics[width=0.6\textwidth]{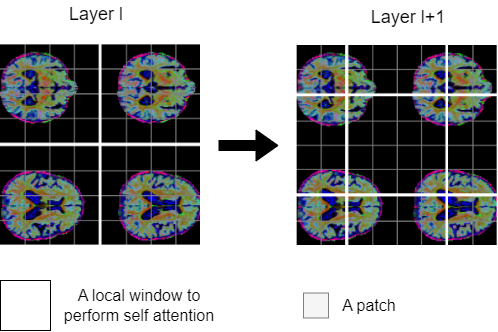}}
\caption{An illustration of the shifted window approach in the Swin Transformer architecture. In the first layer (left), non-overlapping windows are used for self-attention computation within each window. In the subsequent layer (right), the window partitions are shifted, allowing for connections across the boundaries of the previous windows. This shift enables the model to capture dependencies between adjacent regions, improving its ability to model global context while maintaining computational efficiency.}
\label{fig:shiftedwindows}
\end{figure}

The Swin Transformer employs a hierarchical attention mechanism that uses a shifted window approach, as depicted in Fig.~\ref{fig:shiftedwindows}. This approach allows the model to capture both local and global dependencies within the input images. In the shifted window strategy, each self-attention layer operates on a partitioned image, with attention computed locally within non-overlapping windows. In the subsequent layer, the windows are shifted, allowing the model to compute attention across different image regions. This shifting mechanism ensures that the Swin Transformer can model long-range dependencies, which is crucial for processing the spatial relationships in neuroimaging data.

After adapting the Swin Transformer’s input to handle 2D projections of 3D microstructural maps, we fine-tuned the model weights for classification tasks using dMRI metrics. The fine-tuning involved the incorporation of LoRA~\cite{hu2021lora} to refine the attention mechanisms efficiently, and the customization of the output layers for the classification of cognitive impairment categories such as CN, MCI and AD. By integrating LoRA into each attention layer of the Swin Transformer, we adapted the model for early Alzheimer's disease diagnosis while minimizing the number of additional parameters.

The overall architecture, from the generation of microstructural maps to the final classification, is illustrated in Fig.~\ref{fig:swin}.

\subsection{Experiments}

To evaluate the effectiveness of the proposed architecture, we conducted a series of experiments using diffusion MRI data from the ADNI dataset. A data splitting strategy was employed to ensure unbiased evaluation and prevent data leakage. Specifically, 15\% of unique subjects were randomly assigned to a fixed hold-out test set, which was reserved exclusively for final model assessment. The remaining 85\% of subjects were used for 5-fold grouped and stratified cross-validation. This strategy ensured that all data from any given subject appeared entirely within either the training or validation set for a fold, eliminating optimistic bias from intra-subject correlations, while also preserving the original diagnostic class distribution. After cross-validation and hyperparameter tuning, a final model was trained on the combined training and validation data (85\% of the subjects) and used to generate predictions on the held-out test set. To quantify uncertainty, the final reported performance metrics were estimated using 1000 bootstrap samples drawn from the test set predictions and are presented as mean $\pm$ standard error (SE). All data splits were generated using a fixed random seed for reproducibility. 

All subsequent experiments, including model training and inference, were performed on a NVIDIA Tesla A100 GPU to ensure efficient handling of the high-dimensional dMRI data and computational demands of the Swin Transformer. The model was trained using standard backpropagation with the Adam optimizer~\cite{kingma2014adam} and a cosine annealing learning rate schedule~\cite{loshchilov2016sgdr}. The experiments focused on comparing the impact of different microstructural models, analyzing the effect of architectural components via an ablation study, investigating the framework's diagnostic potential for classifying amyloid beta status, and interpreting model decisions through explainable AI.

\subsubsection{dMRI model study}

To assess the impact of different microstructural dMRI models on the performance of the proposed Swin Transformer-based architecture, we conducted experiments utilizing two common dMRI models: DTI and NODDI. These models were chosen to highlight the differences between a single-shell model (DTI) and a multi-shell model (NODDI), reflecting their distinct capabilities in capturing microstructural characteristics relevant to Alzheimer’s disease classification.

The Swin Transformer was trained and evaluated separately on the datasets corresponding to each dMRI model. This allowed us to directly compare the classification performance of the architecture when utilizing DTI-derived metrics versus NODDI-derived metrics. By comparing these models, we aimed to determine which set of diffusion metrics offers better sensitivity and specificity for classifying AD, MCI, and CN subjects.

We evaluated the model's performance on three classification tasks involving different diagnostic groups: (1) distinguishing CN subjects from individuals with AD, (2) differentiating CN from subjects with MCI, and (3) separating MCI from AD cases. The performance metrics (accuracy, precision, recall, and F1 score) are presented in the results section, providing insight into the effectiveness of each dMRI model.

\subsubsection{Ablation study}

To further understand the contributions of each component in our proposed architecture, we conducted an ablation study that systematically evaluates the impact of different architectural choices, including the use of LoRA, and the Swin Transformer's hierarchical attention mechanism.

We designed four main experiments to isolate the effects of these components:

\begin{enumerate}
    \item \textbf{Baseline ResNet}: A 2D ResNet CNN architecture pre-trained on ImageNet was used with the same 2D projections of the dMRI data but without the Swin Transformer or LoRA. This model served as a baseline to compare how the hierarchical attention mechanism and pre-trained weights affect performance.

    \item \textbf{Swin Transformer without LoRA (full fine-tuning)}: In this configuration, we employed the Swin Transformer pre-trained on ImageNet without the incorporation of LoRA. This experiment assesses how the Swin Transformer performs with full-weight fine-tuning on dMRI projections.

    \item \textbf{Swin Transformer without LoRA (head fine-tuning)}: As before, in this configuration, we employed the Swin Transformer pre-trained on ImageNet without the incorporation of LoRA, but only the classification head is fine-tuned. This experiment assesses how the Swin Transformer performs with classification head weight fine-tuning on dMRI projections.

    \item \textbf{Swin Transformer with LoRA}: In this experiment, we applied our proposed architecture with LoRA integrated into the Swin Transformer, evaluating the benefits of LoRA's fine-tuning on the Alzheimer's disease classification task.
\end{enumerate}

Each configuration was trained and evaluated using the same dataset splits to ensure a fair comparison. The performance metrics, including balanced accuracy, precision, recall, and F1 score, were computed for each configuration and compared to understand the individual and combined effects of these components on classification performance.

\subsubsection{Amyloid classification experiments}

To further investigate the diagnostic capabilities of diffusion MRI combined with the Swin Transformer framework, we conducted two experiments focused on classifying amyloid beta status. Amyloid beta is a key biomarker of Alzheimer's disease, often accumulating years before cognitive symptoms appear, making it a valuable target for early detection and intervention in preclinical stages~\cite{villemagne2013amyloid,jack2018nia,sperling2011toward}. Both DTI and NODDI-derived microstructural maps were used to evaluate the discriminatory power of these metrics for amyloid classification.

Amyloid positivity was determined using the UC Berkeley - Amyloid PET 6mm Res analysis derivatives available on the ADNI website. As described in their method, the labels were derived from cortical summary standardized uptake value ratios (SUVRs) calculated from co-registered amyloid PET and MRI data. Subjects with missing amyloid status information were excluded from the analysis.

\paragraph{Experiment 1: Control vs. case classification}

This experiment aimed to distinguish amyloid-negative cognitively normal (A$\beta^-$ CN) subjects from amyloid-positive individuals diagnosed with either MCI or AD (A$\beta^+$ MCI/AD). The amyloid-positive group included 40 MCI and 17 AD subjects and the amyloid-negative group was composed of 107 control subjects. This approach ensures a clearer distinction between healthy control and amyloid-positive cases.

\paragraph{Experiment 2: Amyloid classification in cognitively normal subjects}

The second experiment focused on detecting amyloid beta positivity among CN subjects to assess whether diffusion MRI metrics could distinguish between preclinical amyloid-positive individuals and those without amyloid accumulation. Out of the original ADNI dMRI cohort of 213 CN subjects, 52 individuals were excluded due to missing amyloid status information. The remaining dataset consisted of 161 subjects, categorized into 54 amyloid-positive (A$\beta^{+}$) and 107 amyloid-negative (A$\beta^{-}$) cases.

\subsubsection{Explainability}

To provide insights into the decision-making process of the proposed Swin Transformer-based architecture, we employed Gradient-weighted Class Activation Mapping (Grad-CAM)~\cite{selvaraju2017grad} to visualize which regions of the input contributed most to the model’s predictions. Grad-CAM highlights areas in the input data that are most influential for the classification, which is crucial for enhancing the interpretability of deep learning models in medical imaging.

We applied Grad-CAM to the test set, focusing on correctly classified subjects from each diagnostic group, namely AD, MCI and CN. To obtain a more comprehensive and consistent understanding of the model’s decision-making, we averaged the Grad-CAM output across all correctly classified subjects within each group. This averaging allowed us to identify common patterns and regions of interest (ROI) in the brain that were consistently emphasized by the model during classification. The importance of each region was quantified by calculating the mean value assigned by Grad-CAM to each brain parcellation, based on the Brainnetome atlas (246 ROIs) \cite{fan2016human} and the SUIT cerebellum atlas (34 ROIs) \cite{diedrichsen2006spatially}.

\section{Results}
\label{sec:results}

\subsection{Classification performance}

We evaluated the proposed framework's performance in classifying participants into diagnostic categories using the fine-tuned pre-trained Swin Transformer. We conducted experiments using microstructural maps derived from both DTI and NODDI models. The classification tasks included distinguishing between CN and AD, CN and MCI, and MCI and AD groups.

The classification metrics, including balanced accuracy, precision, recall, and F1 score, are presented in Table~\ref{tab:classification_results}.

\begin{table}[h!]
\sisetup{detect-weight=true}
\centering
\caption{Classification results using Swin Transformer on ADNI data for DTI and NODDI models. All metrics are presented as percentages (mean ± standard error).} 
\label{tab:classification_results}
\begin{tabular}{lcccc}
\hline
 & \textbf{Balanced} & \textbf{Precision} & \textbf{Recall} & \textbf{F1 score} \\
 & \textbf{Accuracy} &  &  &  \\ \hline
\textbf{DTI} & & & &  \\ 
CN vs AD & $91.9\pm3.3$ & $92.3\pm6.5$ & $86.8\pm5.3$ & $88.0\pm4.6$ \\
CN vs MCI & $72.9\pm7.2$ & $75.8\pm6.8$ & $76.1\pm6.3$ & $75.9\pm6.4$ \\
MCI vs AD & $70.7\pm10.3$ & $75.1\pm18.7$ & $74.1\pm8.3$ & $74.5\pm8.2$ \\
 & & & & \\
\textbf{NODDI}  & & & & \\ 
CN vs AD & $95.2\pm2.7$ & $94.5\pm5.3$ & $92.1\pm4.3$ & $92.6\pm3.8$ \\
CN vs MCI & $83.8\pm5.5$ & $84.6\pm7.3$ & $82.6\pm5.5$ & $82.9\pm5.3$ \\
MCI vs AD & $86.8\pm5.1$ & $88.6\pm0.0$ & $81.5\pm7.5$ & $82.3\pm7.0$ \\ 
\hline
\end{tabular}
\end{table}

The results indicate that the model achieved higher classification performance when using NODDI-derived microstructural maps compared to DTI. In particular, the CN vs AD classification task yielded a balanced accuracy of 95.2\% with NODDI, compared to 91.9\% with DTI. Similarly, the CN vs MCI classification showed significant improvement with NODDI, achieving a balanced accuracy of 83.8\%.\\

\subsection{Ablation study}

The ablation study results, presented in Tables~\ref{tab:ablationstudy} and \ref{tab:ablationstudy_MCI} highlight the impact of different model configurations on AD vs CN and MCI vs CN classification tasks. 

For the AD vs CN classification, the baseline ResNet achieved a balanced accuracy of 90.3\% with an F1 score of 85.7\%. Full fine-tuning of the Swin Transformer underperformed with 50\% balanced accuracy, likely due to overfitting. Head fine-tuning improved performance to 91.9\% balanced accuracy and an F1 score of 88.0\%. Adding LoRA further enhanced the model’s performance, achieving the highest balanced accuracy of 95.2\% and an F1 score of 92.6\%.

For the MCI vs CN task, the baseline ResNet showed moderate performance with a balanced accuracy of 75.8\%. Full fine-tuning resulted again in poor performance with a balanced accuracy of 50\%. Head fine-tuning improved results compared to full fine-tuning, reaching 73.9\% of balanced accuracy. The integration of LoRA achieved the best results with a balanced accuracy of 83.8\% and improved recall and F1 scores.

\begin{table}[h!] 
\sisetup{detect-weight=true}
\centering 
\caption{Ablation study: Performance of the model under different settings with the NODDI microstructural model for the \textsc{AD vs CN} classification task. All metrics are presented as percentages (mean ± standard error).}
\label{tab:ablationstudy} 
\begin{tabular}{lcccc} 
\hline 
\textbf{Model Configuration} & \textbf{Balanced} & \textbf{Precision} & \textbf{Recall} & \textbf{F1 score} \\ 
 & \textbf{Accuracy} & & & \\
\hline 
\textbf{Baseline ResNet (2D)} & $90.3 \pm 3.7$ & $91.5 \pm 7.3$ & $84.2 \pm 6.1$ & $85.7 \pm 5.1$ \\
\textbf{Swin Transformer} & $50.0 \pm 0.0$ & $66.6 \pm 0.0$ & $81.6 \pm 6.1$ & $73.3 \pm 8.5$ \\
(full fine-tuning) &  &  &  &  \\
\textbf{Swin Transformer} & $91.9 \pm 3.3$ & $92.3 \pm 6.5$ & $86.8 \pm 5.3$ & $88.0 \pm 4.6$ \\
(head fine-tuning) &  &  &  &  \\
\textbf{Swin Transformer} & $95.2 \pm 2.7$ & $94.5 \pm 5.3$ & $92.1 \pm 4.3$ & $92.6 \pm 3.8$ \\
(LoRA \& head fine-tuning) &  &  &  &  \\
\hline 
\end{tabular} 
\end{table}

\begin{table}[h!] 
\sisetup{detect-weight=true}
\centering 
\caption{Ablation study: Performance of the model under different settings with the NODDI microstructural model for the \textsc{MCI vs CN} classification task. All metrics are presented as percentages (mean ± standard error).}
\label{tab:ablationstudy_MCI} 
\begin{tabular}{lcccc} 
\hline 
\textbf{Model Configuration} & \textbf{Balanced} & \textbf{Precision} & \textbf{Recall} & \textbf{F1 score} \\ 
 & \textbf{Accuracy} & & & \\
\hline 
\textbf{Baseline ResNet (2D)} & $75.8 \pm 6.7$ & $77.5 \pm 7.4$ & $76.1 \pm 6.1$ & $76.5 \pm 6.0$ \\
\textbf{Swin Transformer} & $50.0 \pm 0.0$ & $42.5 \pm 0.0$ & $65.2 \pm 6.8$ & $51.5 \pm 8.6$ \\
(full fine-tuning) &  &  &  &  \\
\textbf{Swin Transformer} & $73.9 \pm 6.5$ & $76.4 \pm 8.5$ & $71.7 \pm 6.5$ & $72.4 \pm 6.3$ \\
(head fine-tuning) &  &  &  &  \\
\textbf{Swin Transformer} & $83.8\pm5.5$ & $84.6\pm7.3$ & $82.6\pm5.5$ & $82.9\pm5.3$ \\
(LoRA \& head fine-tuning) &  &  &  &  \\
\hline 
\end{tabular} 
\end{table}

\subsection{Amyloid classification results}

We evaluated the performance of the Swin Transformer with LoRA fine-tuning in two amyloid classification tasks using DTI- and NODDI-derived microstructural metrics. The model achieved balanced accuracy scores above chance level in both experiments, with NODDI metrics consistently outperforming DTI metrics.

\paragraph{Experiment 1: Control vs. case classification}

In this binary classification task, the model distinguished amyloid-negative cognitively normal (A$\beta^-$ CN) subjects from amyloid-positive cases diagnosed with MCI or AD (A$\beta^+$ MCI/AD). Table~\ref{tab:amyloid_classification_results} summarizes the performance metrics, indicating that NODDI-derived metrics yielded higher accuracy, precision, and recall compared to DTI-derived metrics.

\paragraph{Experiment 2: Amyloid classification in cognitively normal subjects}

In this experiment, we evaluated the model’s ability to classify amyloid status in cognitively normal subjects, identifying individuals with preclinical amyloid accumulation. As shown in Table~\ref{tab:amyloid_classification_results}, NODDI metrics again outperformed DTI metrics.

\begin{table}[h!]
\sisetup{detect-weight=true}
\centering
\caption{Performance metrics for amyloid classification with Swin Transformer on ADNI data for DTI and NODDI models. All metrics are presented as percentages (mean ± standard error).}
\label{tab:amyloid_classification_results}
\begin{tabular}{lcccc
}
\hline
 & \textbf{Balanced} & \textbf{Precision} & \textbf{Recall} & \textbf{F1 score} \\
 & \textbf{Accuracy} &  &  &  \\ \hline
\textbf{DTI} & & & &  \\ 
A$\beta^-$ CN vs A$\beta^+$ MCI/AD & $69.7 \pm 7.0$ & $73.4 \pm 7.5$ & $69.2 \pm 7.6$ & $68.1 \pm 8.1$ \\ 
A$\beta^-$ CN vs A$\beta^+$ CN & $60.7 \pm 11.2$ & $67.4 \pm 12.7$ & $56.5 \pm 10.5$ & $58.0 \pm 10.4$ \\
 & & & & \\
\textbf{NODDI}  & & & & \\ 
A$\beta^-$ CN vs A$\beta^+$ MCI/AD &  $77.2 \pm 6.7$ & $78.9 \pm 7.5$ & $76.9 \pm 6.9$ & $76.6 \pm 7.0$ \\
A$\beta^-$ CN vs A$\beta^+$ CN & $67.9 \pm 9.3$ & $74.9 \pm 12.5$ & $60.9 \pm 10.1$ & $61.9 \pm 10.0$ \\
\hline
\end{tabular}
\end{table}

\subsection{Explainability analysis}

To interpret the model's decisions, we employed Grad-CAM~\cite{selvaraju2017grad} to visualize the regions contributing to the classification. 

\begin{figure}[htbp]
    \centering
    \includegraphics[width=.8\linewidth]{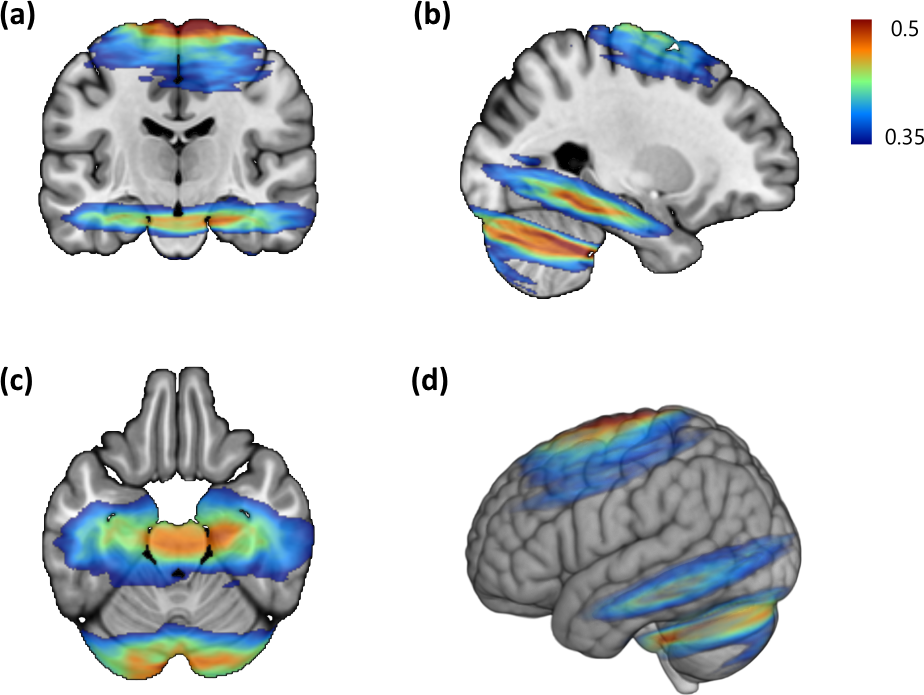}
    \caption{Average activation map generated by Grad-CAM for the subjects correctly classified as AD, highlighting important regions inside the brain contributing to the classification. The regions are represented in 2D (\textbf{(a)} coronal, \textbf{(b)} sagittal and \textbf{(c)} axial) and 3D \textbf{(d)} views in the MNI space.}
    \label{fig:gradcam}
\end{figure}

The activation map resulting from the classification of the CN vs AD groups with the Swin Transformer, enhanced with LoRA and NODDI metrics, is shown in Fig.~\ref{fig:gradcam}. Table~\ref{tab:top_subregions} summarizes the most important brain regions identified by Grad-CAM. Specifically, the table displays the top five most significant \textit{gyri} based on their mean activation values. Within each gyrus, subregions with mean activation values exceeding a threshold of 0.37 were selected, emphasizing the areas that contributed most to the model's classification decisions.

\begin{table}[h!]
\centering
\caption{Top subregions within the five most important \textit{gyri} identified by Grad-CAM, ranked by their mean activation values. Only subregions with mean values greater than 0.37 are displayed.}
\label{tab:top_subregions}
\begin{tabular}{cccc}
\hline
\textbf{Lobe}         & \textbf{Gyrus}                     & \textbf{Subregion}                  & \textbf{Mean} \\
 &  &  & \textbf{Value}  \\
\hline
Temporal Lobe         & Parahippocampal Gyrus (PhG)   & TL (lateral posterior PhG)                   & 0.445  \\
         &   & TH (medial posterior PhG)                    & 0.413  \\
         &    & A28/34 (entorhinal cortex)                   & 0.373  \\
Frontal Lobe          & Superior Frontal Gyrus        & dorsolateral area                            & 0.415  \\
          & Paracentral Lobule            & area 4, (lower limb region)                  & 0.396  \\
Cerebellum            &          /           &                     /               & 0.373  \\
Subcortical Nuclei    & Hippocampus                   & rHipp (rostral hippocampus)                  & 0.379  \\
\hline
\end{tabular}
\end{table}

The analysis revealed that the model attributed more weight to specific brain areas, such as the parahippocampal gyrus (including the entorhinal cortex), cerebellum, hippocampus, superior frontal gyrus, and paracentral lobule.

In contrast, the Grad-CAMs from other classification tasks, with lower balanced accuracy scores, did not reveal any distinct regions of interest.

\section{Discussion}
\label{sec:discussion}

\subsection{Improved classification performance}

Our study demonstrates that leveraging advanced microstructural imaging with the Swin Transformer architecture improves the classification performance for early diagnosis of Alzheimer's disease. The use of NODDI-derived microstructural maps resulted in higher accuracy compared to DTI, suggesting that NODDI provides more sensitive measures of microstructural changes associated with Alzheimer’s disease.

Additionally, our framework was evaluated on amyloid classification tasks, where distinguishing between amyloid-positive and amyloid-negative subjects is critical for early detection of Alzheimer’s disease related pathology. In particular, we examined the classification of cognitively normal subjects based on amyloid positivity. This task targets preclinical stages of Alzheimer’s disease where amyloid deposition is detectable long before the onset of cognitive symptoms. The classification performance for A$\beta^-$ CN vs A$\beta^+$ CN using NODDI-derived metrics yielded promising results, with performance above chance level. This suggests that dMRI metrics, particularly those derived from NODDI, are sensitive enough to detect microstructural changes associated with early amyloid buildup.

\subsection{Insights from the ablation study}

The ablation study offers valuable insights into the contributions of various components in the proposed architecture. First, the baseline 2D ResNet model had inferior performances compared to the Swin Transformer with LoRa, suggesting that the hierarchical attention mechanism and the ability to model global dependencies are useful for capturing diffusion MRI patterns. 

When comparing the different Swin Transformer configurations, the results indicate that fine-tuning the entire model (full fine-tuning) prevents the model from learning, with classification accuracy equivalent to random guessing. This suggests that full fine-tuning is ineffective for smaller datasets, such as the one used in this study, as it may cause the model to overfit or fail to learn meaningful representations.

The best performance was achieved when combining LoRA with head fine-tuning, underscoring the importance of parameter-efficient adaptation in medical imaging tasks, where the availability of labeled data is often limited. LoRA allows for a more targeted fine-tuning approach, updating only a small subset of parameters while maintaining strong generalization. The results also highlight LoRA’s critical role in boosting performance. Without LoRA, head fine-tuning alone resulted in lower balanced accuracy, precision, recall, and F1 scores. By enabling more efficient training with few parameters, LoRA helps to preserve or enhance classification performance, making it a key component in this architecture. This is especially important in neuroimaging applications, where data scarcity can lead to overfitting, and parameter-efficient models are essential to ensure robust learning.

\subsection{Explainability and clinical relevance}

The explainability analysis using the Grad-CAM adapted for our architecture revealed that the model's attention was focused on regions known to be affected in Alzheimer’s disease. This focus was observed in the classification task that achieved the highest balanced accuracy, namely the classification of CN versus AD cases using the Swin Transformer model with LoRA and NODDI metrics. The most prominent regions included the parahippocampal gyrus (including the entorhinal cortex), cerebellum, hippocampus, superior frontal gyrus and paracentral lobule.

These findings align with established neuropathological patterns in Alzheimer’s disease. For example, a review by Rathore et al.~\cite{rathore_review_2017} showed that DTI metrics were significantly impacted by Alzheimer’s disease pathology in key regions such as the hippocampus and posterior cingulate cortex, consistent with our model’s focus areas. This correspondence supports the biological plausibility of our model’s learned features, enhancing its potential clinical relevance.

Furthermore, the regions highlighted by Grad-CAM are consistent with findings from previous studies on the ADNI dataset using CNNs and T1-weighted data, such as Jang and Hwang's work~\cite{jang2022m3t}. Additionally, studies on the ADNI dataset have reported variations in DTI metrics in the temporal and occipital lobes, as well as in the cingulate gyrus and hippocampus, when comparing AD patients to healthy controls~\cite{mueller_ways_2005}. These regions are aligned with our observations, further supporting the model's potential to focus on clinically relevant areas. This alignment highlights the advantages of incorporating explainability features into advanced deep learning architectures.

\subsection{Framework flexibility and adaptability}

The proposed framework leverages pre-trained models and fine-tuning strategies, making it adaptable to other medical imaging applications. The use of 2D projections allows for the utilization of existing pre-trained transformer models, reducing the need for large training datasets. Moreover, the incorporation of LoRA enables efficient fine-tuning with minimal additional parameters, facilitating rapid adaptation to new tasks.

To encourage further research and promote reproducibility, the implementation of our framework is open-source and available on \href{https://github.com/Hyedryn/Swin_dMRI_public}{GitHub}.

\subsection{Comparison with state-of-the-art methods}
\label{sec:sota_comparison}

A notable aspect of this study is the absence of a direct quantitative comparison between our proposed framework and other state-of-the-art methods for Alzheimer's disease or amyloid classification. This decision stems primarily from the distinct modality employed in our work. While numerous deep learning approaches have achieved high performance using structural MRI or functional MRI data~\cite{wen2020convolutional, lianhierarchical2020}, the application of advanced deep learning architectures to multi-shell diffusion MRI data for these specific classification tasks remains underexplored. Direct comparisons across other modalities such as sMRI and fMRI were not included, as these techniques capture fundamentally different biological properties and are not directly interchangeable in terms of input features or clinical interpretation.

Accordingly, this study focused on internal validation through ablation studies and detailed performance reporting specific to dMRI metrics (DTI and NODDI). Furthermore, to facilitate future research and enable direct benchmarking within this specific domain, we have made our implementation open-source and provided the exact subject identifiers used for defining the training, validation, and hold-out test sets. This commitment to transparency and reproducibility aims to establish a reliable baseline against which future methods utilizing multi-shell dMRI for AD and amyloid classification can be directly and fairly compared.

\subsection{Limitations and future work}

Despite the promising results, our study has limitations. The sample size, particularly for the AD group, is relatively small, which may affect the generalizability of the findings. Future work should include larger and more diverse cohorts to validate the model's robustness. Additionally, while the use of 2D projections allows the framework to leverage pre-training on large-scale datasets like ImageNet, it may not fully capture the intricate 3D spatial relationships inherent in dMRI data. Extending the framework to process 3D volumes directly could potentially improve performance by preserving spatial context, but this would come at the cost of losing the advantages of pre-training and may require significantly larger datasets to train effectively from scratch. Balancing these trade-offs will be an important direction for future research.

Another limitation in our amyloid analysis relates to the sample size and dataset composition. While we compared A$\beta^-$ CN versus A$\beta^+$ CN and A$\beta^-$ CN versus A$\beta^+$ MCI/AD, the relatively small number of individuals with both multi-shell dMRI and amyloid status in the ADNI database limited the statistical power of these comparisons. This constraint may affect the robustness of the findings and restrict the exploration of amyloid's role in more granular diagnostic stratifications. However, as the ADNI database continues to grow, the availability of subjects with comprehensive imaging and biomarker data is expected to increase, enabling larger and more representative analyses.

Furthermore, tau PET imaging data are becoming increasingly available in the ADNI database. As tau is a critical biomarker of Alzheimer's disease alongside amyloid accumulation, incorporating both biomarkers could refine classification tasks and improve the biological interpretability of models. Future studies could integrate these complementary biomarkers to stratify subjects into more meaningful categories, enhancing the granularity and precision of diagnostic tools. These advances, coupled with larger and more diverse datasets, will also facilitate the training of more sophisticated multi-modal deep learning frameworks.

\section{Concluding remarks}

Our study demonstrates the potential of combining a pre-trained Swin Transformer architecture with advanced microstructural models for accurate classification using diffusion MRI. By leveraging diffusion models such as NODDI and DTI, along with parameter-efficient techniques like LoRA, we were able to capture subtle microstructural changes for neuroimaging-based classification tasks. The results underline LoRA's contribution to enhancing classification accuracy, particularly in data-constrained scenarios, by optimizing model adaptation without extensive retraining. As diffusion MRI continues to evolve as a critical tool for probing tissue microstructure, the presented methods hold significant promise for advancing research applications in neuroimaging.

\section*{Funding}
Quentin Dessain is a research fellow of the Fonds de la Recherche Scientifique - FNRS of Belgium. Nicolas Delinte was funded by the MedReSyst-AI4Alzheimer project, which has been supported by the European Union and Wallonia as part of the "Wallonia 2021-2027" program. Bernard Hanseeuw is a clinical researcher of FNRS, supported by the Wel Research Institute (WelBIO research program) and the StopAlzheimer Foundation. Computational resources have been provided by the Consortium des Équipements de Calcul Intensif (CÉCI), funded by the Fonds de la Recherche Scientifique de Belgique (F.R.S.-FNRS) under Grant No. 2.5020.11 and by the Walloon Region. Data collection and sharing for this project was funded by the Alzheimer's Disease Neuroimaging Initiative (ADNI) (National Institutes of Health Grant U01 AG024904).

\section*{Data availability statement}

The code supporting the findings of this study are available in the GitHub repository of the article at \url{https://github.com/Hyedryn/Swin_dMRI_public}.

\bibliographystyle{ieeetr}
\bibliography{bib}

\end{document}